\documentclass[10pt,twocolumn]{article}
\usepackage[letterpaper,top=0.72in,bottom=0.75in,left=0.69in,right=0.69in]{geometry}
\usepackage{fontspec}
\newfontfamily\nepalifont{Mukta-Regular.ttf}[Script=Devanagari]
\newcommand{\dn}[1]{{\nepalifont #1}}
\usepackage{amsmath,amssymb}
\usepackage{graphicx}
\usepackage{booktabs,array,multirow,tabularx}
\usepackage{xurl}
\usepackage{microtype}
\usepackage{flushend}
\usepackage{cuted}
\usepackage{caption}
\usepackage{xcolor}
\definecolor{resourceblue}{RGB}{16,76,137}
\usepackage[colorlinks=true,urlcolor=resourceblue,linkcolor=black,citecolor=black]{hyperref}
\usepackage{titlesec}
\titleformat{\section}{\normalfont\bfseries\large}{\thesection.}{0.6em}{}
\titleformat{\subsection}{\normalfont\bfseries\normalsize}{\thesubsection}{0.6em}{}
\titlespacing*{\section}{0pt}{12pt}{5pt}
\titlespacing*{\subsection}{0pt}{9pt}{3pt}
\newcommand{\TableOne}{%
\begin{table}[!t]\centering
\caption{Source clips whose transcript contains a character the SraVaani tokenizer cannot encode, by cause.}\label{tab:unencodable}
\resizebox{\columnwidth}{!}{\begin{tabular}{lrrrr}\toprule
Subset & Clips & Unenc. \% & Digits \% & Punct. \%\\\midrule
OpenSLR Nepali & 72,283 & 7.5 & 7.4 & 0.1\\
Lectures & 20,138 & 86.3 & 38.9 & 47.3\\
Podcast & 7,165 & 62.7 & 24.1 & 38.6\\
News \& talk & 14,820 & 42.2 & 14.4 & 27.8\\
Conversation & 997 & 22.6 & 8.1 & 14.4\\
Interviews & 772 & 93.5 & 9.2 & 84.3\\\midrule
All & 116,175 & 29.7 & 14.8 & 14.9\\\bottomrule
\end{tabular}}\end{table}}

\newcommand{\TableTwo}{%
\begin{table}[!t]\centering
\caption{Spelling out numbers with the base model's output as evidence (excerpts). The last three examples are from engineering lectures.}\label{tab:numberexamples}
\resizebox{\columnwidth}{!}{\begin{tabular}{p{3.0cm}p{3.4cm}p{3.0cm}}\toprule
Original & Base-model hypothesis & Final label\\\midrule
1947 \dn{मा} India \dn{मा} & \dn{उन्नाइस सय सय चालिसमा भारतमा} & \dn{उन्नाइस सय सतचालीस मा} india \dn{मा}\\
\dn{त्यसलाई} 1 \dn{सय} 50 degree & \dn{त्यसलाई एक सय पचास डिग्री} & \dn{त्यसलाई एक सय पचास} degree\\
\dn{अब जस्तै} 32 27 \dn{होइन} 24 24 & \dn{अब जस्तै बत्तीस सत्ताइस होइन चौबीस चौबीस} & \dn{अब जस्तै बत्तीस सत्ताइस होइन चौबीस चौबीस}\\
6 into 10 \dn{को} power & \dn{सिक्स इन्टो टेनको पावर} & six into ten \dn{को} power\\
FXK+1 & \dn{एफ एक्स के प्लस वान} & f x k plus one\\
T.remove teacher & \dn{टी डट रिमुभ टिचर} & t dot remove teacher\\\bottomrule
\end{tabular}}\end{table}}

\newcommand{\TableThree}{%
\begin{table*}[!t]\centering
\caption{The final corpus by subset and split. CM: share of clips with at least one English word; EN: share of English words among all words.}\label{tab:corpus}
\resizebox{0.94\textwidth}{!}{\begin{tabular}{llrrrrrrrr}\toprule
& & \multicolumn{2}{c}{Train} & \multicolumn{2}{c}{Validation} & \multicolumn{2}{c}{Test} & & \\
Subset & Domain & Clips & h & Clips & h & Clips & h & CM \% & EN \%\\\midrule
OpenSLR Nepali & read speech & 69,133 & 57.62 & 1,472 & 1.22 & 1,500 & 1.25 & 15.1 & 6.6\\
Lectures & engineering/CS lectures & 15,867 & 44.53 & 2,277 & 6.35 & 1,574 & 4.25 & 98.4 & 35.9\\
Podcast & podcast / talk show & 6,305 & 28.47 & 363 & 1.63 & 358 & 1.63 & 85.1 & 13.1\\
News \& talk & news and conversation & 12,990 & 19.61 & 816 & 1.29 & 730 & 1.12 & 85.9 & 26.3\\
Conversation & code-mixed conversation & 570 & 0.99 & 48 & 0.08 & 373 & 0.62 & 95.5 & 36.2\\
Interviews & technical interviews & 583 & 1.68 & 73 & 0.20 & 75 & 0.21 & 100.0 & 47.0\\\midrule
\textbf{Total} & & \textbf{105,448} & \textbf{152.91} & \textbf{5,049} & \textbf{10.77} & \textbf{4,610} & \textbf{9.08} & \textbf{43.8} & \textbf{24.2}\\\bottomrule
\end{tabular}}\end{table*}}

\newcommand{\TableFour}{%
\begin{table}[!t]\centering
\caption{Number of clips removed at each preparation step.}\label{tab:removals}
\resizebox{\columnwidth}{!}{\begin{tabular}{lrrr}\toprule
& Train & Val & Test\\\midrule
Source clips & 106,347 & 5,138 & 4,690\\
\quad Ambiguous number reading (tie) & 519 & 66 & 59\\
\quad Hypothesis too far (CER $>0.5$) & 287 & 15 & 13\\
\quad Label too long for CTC & 93 & 8 & 8\\\midrule
Final clips & 105,448 & 5,049 & 4,610\\
Final hours & 152.9 & 10.8 & 9.1\\\bottomrule
\end{tabular}}\end{table}}

\newcommand{\TableFive}{%
\begin{table}[!t]\centering
\caption{SraVaani components and parameter counts, from the released checkpoint. All parameters are fine-tuned in our main configuration.}\label{tab:components}
\resizebox{\columnwidth}{!}{\begin{tabular}{llr}\toprule
Component & Structure & Params\\\midrule
Front end & log-mel, 128 bins (fixed) & --\\
Subsampling & 3 strided conv. stages, 256 ch. + linear & 4.3M\\
Conformer blocks & $17\times$ ($d=1024$, 8 heads, kernel 9) & 428.6M\\
Prediction network & embedding + LSTM, 640 units & 6.5M\\
Joint network & $1024+640\rightarrow 640\rightarrow 5006$ & 4.3M\\
CTC head & $1024\rightarrow 5001$ & 5.1M\\\midrule
Total & & 448.8M\\\bottomrule
\end{tabular}}\end{table}}

\newcommand{\TableSix}{%
\begin{table}[!t]\centering
\caption{Our fine-tuning settings compared with the official SraVaani fine-tuning script.}\label{tab:settings}
\resizebox{\columnwidth}{!}{\begin{tabular}{lll}\toprule
Setting & Official script & Ours\\\midrule
Encoder & frozen & trained, LR $\times 0.5$\\
Peak LR (decoders) & $10^{-4}$ & $10^{-4}$\\
Warm-up & 10 updates & 5\% (494 updates)\\
Schedule & cosine to $10^{-6}$ & cosine to $10^{-6}$\\
Clips per update & $4\times 8=32$ & $8\times 8=64$\\
Epochs & 2 & 6 (and 12)\\
Precision & fp32 & bf16 mixed\\
Joint sub-batch & 4 & 1\\
Validation & once per epoch & twice per epoch\\
Checkpoint selection & last weights & best validation WER\\\bottomrule
\end{tabular}}\end{table}}

\newcommand{\TableSeven}{%
\begin{table}[!t]\centering
\caption{FLEURS reference adaptation to the mixed-script convention.}\label{tab:fleursrefs}
\resizebox{\columnwidth}{!}{\begin{tabular}{p{5.5cm}p{5.5cm}}\toprule
Original reference & Mixed-script reference\\\midrule
\dn{टेलिभिजन रिपोर्टहरूले प्लान्टबाट सेतो धुवाँ आउँदै गरेको देखाउँछन्} & television report \dn{हरूले} plant \dn{बाट सेतो धुवाँ आउँदै गरेको देखाउँछन्}\\
\dn{गिजामा रहेको ग्रेट पिरामिड सातवटा आश्चर्यमध्ये एउटा मात्रै हो जुन अहिले पनि उभिरहेको छ} & giza \dn{मा रहेको} great pyramid \dn{सातवटा आश्चर्यमध्ये एउटा मात्रै हो जुन अहिले पनि उभिरहेको छ}\\
\dn{हट चकलेट बेल्जियमको मानक स्तरमा छ फलका रस महङ्गो भए पनि उत्कृष्ट छ} & hot chocolate \dn{बेल्जियमको मानक स्तरमा छ फलका रस महङ्गो भए पनि उत्कृष्ट छ}\\\bottomrule
\end{tabular}}\end{table}}

\newcommand{\TableEight}{%
\begin{table*}[!t]\centering
\caption{Test results of all models (\%). Rom-WER ignores script. FLEURS is out of domain. All fine-tuned models use the same training data; bold marks the best fine-tuned result in each column.}\label{tab:results}
\resizebox{\textwidth}{!}{\begin{tabular}{lrrrrrrrrrrrrr}\toprule
Model & Params & WER & CER & EN-WER & NE-WER & Rom-WER & OpenSLR & Lect. & Podc. & News & Conv. & Interv. & FLEURS\\\midrule
\multicolumn{14}{l}{\emph{Zero-shot}}\\
SraVaani & 449M & 48.6 & 33.3 & 80.4 & 36.1 & 41.3 & 62.5 & 43.7 & 47.1 & 55.7 & 54.9 & 62.6 & 22.6\\
Whisper large-v3-turbo & 809M & 121.5 & 81.6 & 86.0 & 135.4 & 115.4 & 106.1 & 113.2 & 126.9 & 156.7 & 110.0 & 98.3 & 88.4\\
IndicConformer & 120M & 49.1 & 37.2 & 100.0 & 29.3 & 41.6 & 16.3 & 51.5 & 48.0 & 53.0 & 54.4 & 61.5 & 22.2\\\midrule
\multicolumn{14}{l}{\emph{Fine-tuned on our training data}}\\
Whisper large-v3-turbo & 809M & 16.4 & 7.5 & \textbf{14.8} & 17.0 & 14.8 & 16.8 & 14.0 & 20.3 & 20.9 & \textbf{12.8} & \textbf{11.6} & 36.9\\
IndicConformer & 128M & 19.9 & 11.3 & 26.5 & 17.4 & 18.4 & \textbf{12.0} & 17.8 & 23.5 & 25.0 & 21.3 & 18.6 & 32.2\\
SraVaani, 6 epochs & 449M & 17.1 & 8.0 & 20.1 & 15.9 & 15.5 & 12.4 & 15.4 & 20.9 & 21.0 & 14.7 & 13.9 & \textbf{31.1}\\
SraVaani, 12 epochs & 449M & \textbf{16.3} & \textbf{7.5} & 18.8 & \textbf{15.3} & \textbf{14.7} & 12.1 & 14.8 & \textbf{20.1} & \textbf{19.0} & 14.7 & 12.8 & 31.4\\\bottomrule
\end{tabular}}\end{table*}}

\newcommand{\TableNine}{%
\begin{table*}[!t]\centering
\caption{Errors by language on the test set (\%). CM / Mono: WER on clips with and without English words. Wrong script: English words written in Devanagari. Switch / Int.: error rate next to a language switch and elsewhere. Section~\ref{sec:metrics} defines the metrics.}\label{tab:language}
\resizebox{0.87\textwidth}{!}{\begin{tabular}{lrrrrrrrrr}\toprule
System & CM & Mono & EN-WER & NE-WER & EN exact $\uparrow$ & Wrong script & Rom-WER & Switch & Int.\\\midrule
Zero-shot & 49.6 & 42.1 & 80.4 & 36.1 & 21.2 & 11.6 & 41.3 & 65.9 & 35.5\\
Fine-tuned, 6 epochs & 17.4 & 15.2 & 20.1 & 15.9 & 82.3 & 0.3 & 15.5 & 14.3 & 15.2\\
Fine-tuned, 12 epochs & 16.6 & 14.4 & 18.8 & 15.3 & 83.6 & 0.3 & 14.7 & 13.6 & 14.4\\\bottomrule
\end{tabular}}\end{table*}}

\newcommand{\TableTen}{%
\begin{table}[!t]\centering
\caption{Ablations on the test set (WER, \%). $\Delta$: change from the main configuration, with its 95\% confidence interval. Num./Other: clips with and without numbers in the source transcript.}\label{tab:ablations}
\resizebox{\columnwidth}{!}{\begin{tabular}{lrlrrrr}\toprule
System & WER & $\Delta$ [95\% CI] & EN-WER & Num. & Other & FLEURS\\\midrule
Main (full fine-tune) & 17.1 & -- & 20.1 & 14.3 & 18.7 & 31.1\\
\quad seed 2 & 17.0 & $-0.1$ [$-0.3,+0.1$] & 19.7 & 14.4 & 18.5 & 31.7\\
\quad 12 epochs & 16.3 & $-0.8$ [$-1.0,-0.6$] & 18.8 & 13.6 & 17.8 & 31.4\\
\quad CTC decoding & 19.1 & $+2.0$ [$+1.8,+2.2$] & 23.8 & 16.1 & 20.8 & 32.4\\
Frozen encoder & 25.5 & $+8.4$ [$+8.0,+8.8$] & 30.4 & 22.0 & 27.5 & 31.8\\
No numeric clips & 19.8 & $+2.7$ [$+2.4,+3.0$] & 22.3 & 21.3 & 19.0 & 31.8\\
No OpenSLR & 18.5 & $+1.4$ [$+1.1,+1.6$] & 20.2 & 14.9 & 20.5 & 36.2\\\bottomrule
\end{tabular}}\end{table}}

\newcommand{\TableEleven}{%
\begin{table}[!t]\centering
\caption{How often number words in the test references are recognised wrongly (substitutions and deletions, \%), by the language they were spoken in.}\label{tab:numbererrors}
\resizebox{\columnwidth}{!}{\begin{tabular}{lrr}\toprule
System & English (1,617 words) & Nepali (3,744 words)\\\midrule
Zero-shot & 88.0 & 32.6\\
No numeric clips & 16.6 & 29.8\\
Fine-tuned, 6 epochs & 4.9 & 6.4\\\bottomrule
\end{tabular}}\end{table}}

\newcommand{\TableTwelve}{%
\begin{table}[!t]\centering
\caption{Results on FLEURS Nepali (\%). WER$_\mathrm{CM}$: WER against references rewritten in our code-mixed convention (Sec.~\ref{sec:fleurs-results}). Rom-CER$_{\neg\mathrm{sp}}$ ignores script and spaces; Latin: share of output words in Latin script.}\label{tab:fleurs}
\resizebox{\columnwidth}{!}{\begin{tabular}{lrrrrrr}\toprule
System & WER & WER$_\mathrm{CM}$ & CER & Rom-WER & Rom-CER$_{\neg\mathrm{sp}}$ & Latin\\\midrule
SraVaani zero-shot & 22.6 & 30.0 & 6.8 & 17.8 & 5.1 & 0.0\\
SraVaani fine-tune, 6 epochs & 31.1 & 23.4 & 15.6 & 26.9 & 7.6 & 10.1\\
SraVaani fine-tune, 12 epochs & 31.4 & 24.0 & 15.4 & 27.1 & 7.6 & 9.8\\
Whisper zero-shot & 88.4 & 87.9 & 28.0 & 74.9 & 18.0 & 0.4\\
Whisper fine-tune & 36.9 & 29.0 & 16.8 & 30.1 & 8.4 & 9.1\\
IndicConformer zero-shot & 22.2 & 29.1 & 6.8 & 17.5 & 5.1 & 0.0\\
IndicConformer fine-tune & 32.2 & 26.4 & 16.3 & 27.6 & 9.2 & 9.0\\\bottomrule
\end{tabular}}\end{table}}

\newcommand{\TableThirteen}{%
\begin{table*}[!t]
\centering
\caption{Representative transcript examples. The fine-tuned model usually retains English in Latin script. The final example illustrates a remaining error.}\label{tab:examples}
\begingroup\small
\setlength{\tabcolsep}{4pt}\renewcommand{\arraystretch}{1.23}
\begin{tabularx}{\textwidth}{@{}p{1.76cm}>{\raggedright\arraybackslash}X>{\raggedright\arraybackslash}X>{\raggedright\arraybackslash}X@{}}
\toprule
Domain & Reference & Zero-shot & Fine-tuned, 6 epochs\\\midrule
Lectures & a one a one dot length a one \dn{को} length \dn{सम्म} & \dn{ए वान कमा ए वान डट लेन्थ ए वानको लेन्थसम्म} & a one a one dot length a one \dn{को} length \dn{सम्म}\\
Conversation & responsibility \dn{दिँदा} project after \dn{पनि} how they will continue \dn{भन्ने कुरा छ} & \dn{रेस्पोन्सिबिलिटी दिँदा प्रोजेक्ट आफ्टर पनि हाउ दे विल कन्टिन्यू भन्ने कुरा छ} & responsibility \dn{दिँदा} project after \dn{पनि} how they will continue \dn{भन्ने कुरा छ}\\
Lectures & \dn{यो भनेको} zero point three five x \dn{भनेको} thirty five \dn{भयो} & \dn{यो भनेको जिरो पोइन्ट थ्री फाइभ एक्स भनेको थर्टी फाइभ भयो} & \dn{यो भनेको} zero point three five x \dn{भनेको} thirty five \dn{भयो}\\
Conversation & certificate \dn{पायो भने} we are very very lucky & \dn{सर्टिफिकेट पाय भने वी आर वेरी वेरी लकी} & certificate \dn{पायो भने} we are very very lucky\\
Lectures & \dn{कति अरू पढियो} girls fast student teacher doctor master & golf fast student teacher doctor master & \dn{केटीहरुले बाँकी गर्नुहोस्} gulf fast student teacher doctor master \dn{ए}\\
\bottomrule
\end{tabularx}
\endgroup
\end{table*}}

\newcommand{\FigureOne}{%
\begin{figure}[!t]\centering
\includegraphics[width=\columnwidth]{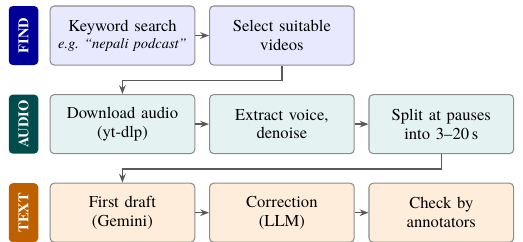}
\caption{YouTube speech collection: video selection, audio processing, segmentation and transcript verification.}\label{fig:collection}\end{figure}}
\newcommand{\FigureTwo}{%
\begin{figure}[!t]\centering
\includegraphics[width=\columnwidth]{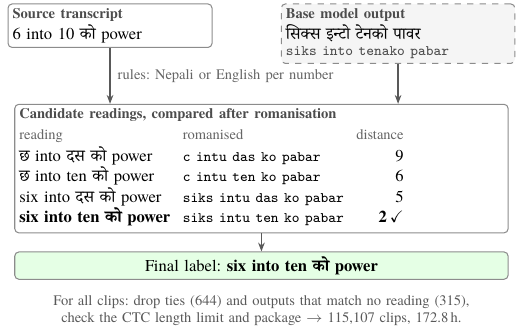}
\caption{Spelling out numbers, on a real lecture clip. Rules propose a Nepali and an English reading for each number. We romanise the candidates and the base model's output, which writes English in Devanagari, and keep the closest candidate.}\label{fig:verbalisation}\end{figure}}
\newcommand{\FigureThree}{%
\begin{figure}[!t]\centering
\includegraphics[width=\columnwidth]{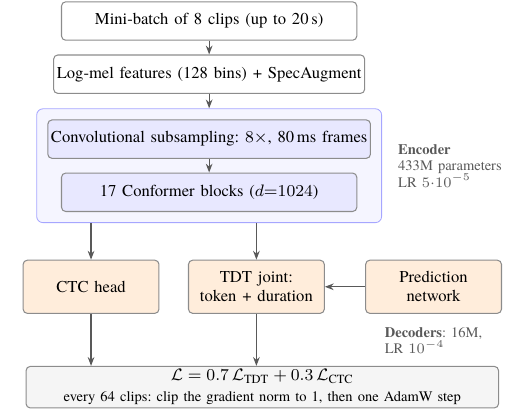}
\caption{One fine-tuning step of SraVaani. The two decoders share the encoder and the 5,000-piece vocabulary. All 449M parameters are trained, and the encoder gets half the learning rate of the decoders.}\label{fig:training}\end{figure}}
\newcommand{\FigureFour}{%
\begin{figure}[!t]\centering
\includegraphics[width=\columnwidth]{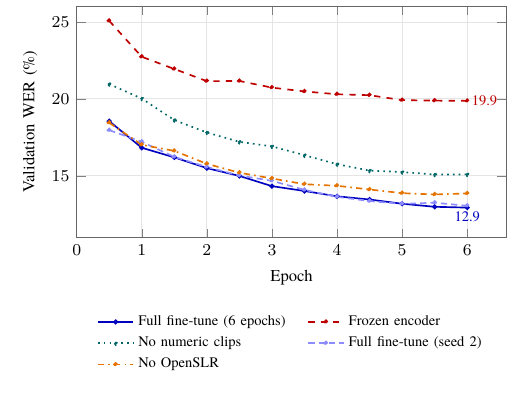}
\caption{Six-epoch validation WER for the main configuration and ablations. Validation is performed twice per epoch.}\label{fig:ablation-curves}\end{figure}}
\newcommand{\FigureFive}{%
\begin{figure}[!t]\centering
\includegraphics[width=\columnwidth]{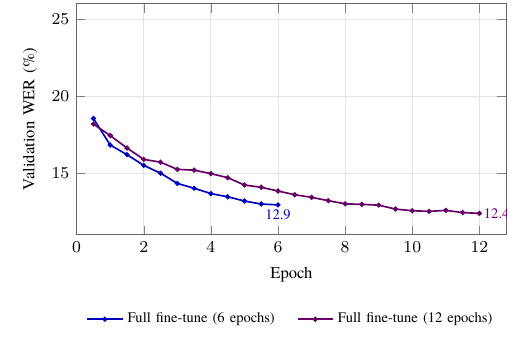}
\caption{Validation WER for six- and twelve-epoch full fine-tuning. Each run uses a cosine schedule matched to its training duration. Validation is performed twice per epoch.}\label{fig:training-curves}\end{figure}}

\begin{document}
\title{Nepali--English Code-Mixed Speech Recognition:\\Data Preparation and Adaptation of a Speech Foundation Model}
\author{Jeevan Neupane, Janardan Bhetwal, Kshitiz Paudel, Anup Aryal, Bipin Bashyal\\
\href{https://ailab.com.np/}{AI Lab Pvt. Ltd.}, Nepal\\[-1pt]
{\footnotesize\texttt{\{jeevanneupane, janardanbhetwal, kshitizpaudel, anuparyal, bipinbashyal\}@ailab.com.np}}}
\date{}
\maketitle
\par\medskip\noindent\textbf{Abstract}\par\smallskip
Nepali speakers regularly mix English words into conversation, but multilingual speech recognisers often struggle with language switches and the writing script used for English words. We study how to adapt SraVaani, a 449M-parameter hybrid TDT--CTC model, using a 172.8-hour corpus covering six Nepali speech domains. Its pretrained tokenizer does not support digits and several common symbols. Rather than replace it or discard these recordings, we generate Nepali and English readings for numbers and use zero-shot ASR output to select a plausible form. Full fine-tuning reduces test word error rate (WER) from 48.6\% to 17.1\%; a 12-epoch run reaches 16.3\%, compared with 16.4\% for fine-tuned Whisper large-v3-turbo. Freezing the encoder increases WER to 25.5\%, while excluding numeric clips raises WER on numeric test speech from 14.3\% to 21.3\%. We also evaluate on FLEURS using both its original references and mixed-script alternatives. The comparison shows that transcription conventions explain part, but not all, of the out-of-domain performance loss.

\section{Introduction}
Nepali speech frequently includes English words, particularly in discussions of education, work and technology. Speakers switch between languages within a sentence, but the written transcript is expected to retain Nepali in Devanagari and English in Latin script. A model can recognise the spoken word yet still produce the wrong script. Under standard word error rate (WER), that difference is treated as a recognition error.

Large multilingual recognisers such as Whisper~\cite{ref1} and Indic models trained on broad speech collections~\cite{ref2,ref3,ref4} provide a useful starting point. They do not, however, make the data preparation problem disappear. SraVaani was trained mainly on single-script utterances. On our evaluation set, its zero-shot output places many English words in Devanagari and sometimes selects another Indic script for short Nepali speech. The training labels also contain digits and punctuation that its fixed tokenizer cannot represent.

Numbers are especially difficult. The written form ``1947'' may correspond to a Nepali or English reading, while technical recordings contain equations and programming expressions. Removing these clips loses relevant speech. Replacing every digit with one language's number words can produce labels that do not match the audio. We therefore construct alternative readings and select among them using the base model's output as evidence, followed by quality checks.

Our study addresses three practical questions: how to prepare mixed-script transcripts without rebuilding the tokenizer, whether full encoder adaptation and numeric clips improve recognition, and how writing conventions affect reported error rates. We prepare 172.8 hours across six domains, compare full and restricted fine-tuning, evaluate against Whisper and IndicConformer, and test the same out-of-domain FLEURS speech under two reference conventions. These comparisons are intended to identify which changes matter, not to establish a new general-purpose state of the art.

\section{Related Work}
\noindent\textbf{Multilingual and Indic ASR.} Whisper~\cite{ref1} and MMS~\cite{ref5} cover Nepali, while IndicVoices~\cite{ref2}, Vistaar~\cite{ref7} and VAANI~\cite{ref4} have expanded Indic speech training and evaluation. SraVaani~\cite{ref3} combines a FastConformer encoder~\cite{ref9} with Token-and-Duration Transducer (TDT) and CTC outputs~\cite{ref8}. We retain its pretrained vocabulary and both output heads.

\noindent\textbf{Code-mixed recognition and scoring.} The MUCS challenge~\cite{ref11} established Hindi--English and Bengali--English ASR benchmarks, and earlier work discusses code-switching and monolingual forgetting~\cite{ref10,ref13}. Transliteration-aware measures~\cite{ref16,ref18}, language-specific metrics such as PIER~\cite{ref20} and switch-point evaluation~\cite{ref21} show why a single WER value can hide different errors. For Nepali, Devkota~\cite{ref36} studied code-mixed transcription and script conversion, while Paudel~\cite{ref37} examined spontaneous Nepali--English podcast speech. Other Nepali comparisons include Whisper adaptation~\cite{ref32} and multi-model evaluation~\cite{ref33}. We differ in our emphasis on fixed-tokenizer preparation, bilingual numbers and encoder adaptation.

\noindent\textbf{Normalisation and evaluation corpora.} Audio-guided text normalisation has been explored with NeMo~\cite{ref25}. We use a related selection principle but generate candidates in both languages and compare approximate pronunciation rather than raw script. OpenSLR Nepali~\cite{ref29} supplies read speech, and FLEURS~\cite{ref34} provides a public out-of-domain reference. Our mixed-script FLEURS version tests the influence of transcription style without changing its audio.

\section{Dataset Collection and Preparation}
We combine public Nepali recordings with speech collected from online material. The preparation pipeline covers audio cleaning, transcript checks, train/test separation and compatibility with SraVaani's fixed tokenizer. Figures~\ref{fig:collection} and~\ref{fig:verbalisation} outline the main steps.
\FigureOne

\subsection{Speech sources}
The corpus draws on six speech settings (Table~\ref{tab:corpus}). OpenSLR supplies mostly read Nepali sentences. The remaining sources include engineering lectures, podcasts, news and talk programmes, conversations and technical interviews. They differ in speaking style, background noise and the amount of English. Lectures, for instance, contain programming terms and mathematical expressions that are uncommon in ordinary read speech.

\subsection{Collection of in-the-wild speech}
For the YouTube material, annotators first searched for suitable Nepali videos using a small tool built on the YouTube Data API. Audio was downloaded with \texttt{yt-dlp}. We separated speech from background sounds, used a 300 Hz--8 kHz band-pass filter and applied denoising. Voice activity detection then split recordings at pauses of at least 500 ms, targeting chunks of 3--20 seconds.

Gemini produced the first transcripts. A second LLM pass corrected them, and human annotators checked the text against the recordings. The transcription convention was Nepali in Devanagari and English words in Latin script. The 3--20 second target describes this collection step; clips from the other sources could be longer.

\subsection{Audio and transcript quality control}
We converted audio to 16 kHz, mono, 16-bit PCM. MMS forced alignment~\cite{ref5} helped trim speech at clip boundaries. We removed clips with poor transcript coverage and checked for duplicate audio across subsets. We also removed train-set sentences that were duplicated in held-out splits, where these could be identified.

A second automatic transcript helped flag possible mistakes for review. We excluded 604 training clips during the earlier quality filtering step: 377 with large disagreement between two transcripts, 92 with repetition artefacts, 98 duplicate held-out sentences and 41 with implausible speech rates (four clips met two criteria). This left 116,175 source clips or 175.2 hours before tokenizer-specific cleaning. The existing train, validation and test assignments were kept.

\subsection{Tokenizer compatibility}
SraVaani has a 5,000-piece SentencePiece BPE vocabulary~\cite{ref38}. It can represent Nepali and English letters, but not common digits or symbols such as the percent sign, punctuation marks and apostrophes. Unsupported characters are mapped to the unknown token. Table~\ref{tab:unencodable} shows that 29.7\% of source clips contain at least one such character; 14.8\% contain digits and 14.9\% contain punctuation or symbols without digits.

We retain the original vocabulary to reuse the pretrained prediction network and output heads. Instead, we changed the transcripts to valid strings. This preserves those model weights, although English words may be divided into several short token pieces.
\TableOne

\subsection{Transcription convention}
The training targets follow four rules: Nepali words use Devanagari; English words use lowercase Latin script; punctuation is removed; and numbers and unsupported symbols are written as spoken words. Hyphens become spaces, apostrophes are removed, and we apply Unicode NFKC normalisation followed by single spacing. We check every final label by encoding and decoding it with the original tokenizer. A label is accepted only when no unknown token appears and the round-trip text is unchanged.

\subsection{Bilingual number verbalisation}
Digits alone do not show which language the speaker used. The string ``1947'' may be read in Nepali or English, and engineering recordings contain forms such as ``6 into 10'', ``FXK+1'' and program identifiers with dots. We generate possible Nepali and English readings instead of applying a single rule to each number.

The rule-based verbaliser covers more than 40 patterns, including integers, years, decimals, ranges, percentages, ordinals, times, fractions, superscripts, subscripts, alphanumeric forms and code expressions. It builds candidate utterances from Nepali and English number words, keeping at most 32 complete-sentence candidates per clip.

Next, we run the base SraVaani model on the audio. Its output often writes English sounds in Devanagari. To compare outputs fairly, we map the text to a rough romanised pronunciation key, merging common spelling differences. We score the candidate strings by Levenshtein distance to this key. After choosing a sentence-level candidate, we also try alternative readings for individual numbers. Figure~\ref{fig:verbalisation} shows one example, and Table~\ref{tab:numberexamples} gives more cases.
\FigureTwo
\TableTwo

We reject clips when the two best readings tie (644 clips) or when the closest candidate is still too far from the model output (315 clips, using a normalised distance threshold of 0.5). Among 35,157 resolved numbers, 45.1\% use Nepali readings and 53.1\% use English readings. The distribution varies by source: 99.0\% of numbers in OpenSLR are read in Nepali, while 75.6\% of numbers in engineering lectures are read in English.

We manually audited 150 numeric clips sampled across the sources. Five contained errors (3.3\%; 95\% confidence interval 1.4--7.6\%). Several errors came from missing reading patterns, such as an extra \dn{एक} before hundreds or an operator that was not spoken. Adding these possibilities changed the chosen readings in 539 of 634 affected clips according to the same automatic scoring method. A separate sample of 100 clips had no identified errors in nonnumeric normalisation. The corrected numeric labels still need independent listening-based verification.

\subsection{Length checks and data packaging}
SraVaani's encoder emits one frame every 80 ms. With $T=\lfloor d/0.08\rfloor$ frames for a clip of $d$ seconds, a CTC label of $U$ tokens and $R$ adjacent repeated tokens must satisfy
\begin{equation}
U\leq T-2,\qquad U+R\leq T.
\end{equation}
This check removes 109 clips. We also cap training clips at 20 seconds, which causes the data loader to skip another 33 clips during training; validation and test retain their longer clips.

We store the audio and manifests in NeMo's tarred format~\cite{ref40}, using 32 training shards and two shards each for validation and test. The audio bytes and the manifest entries have matching orders. We check file duration, unique clip IDs, absence of duplicate audio hashes across splits, tokenizer round trips and full manifest coverage by the loader.

\subsection{Corpus statistics}
The final dataset has 115,107 clips and 172.8 hours of audio. Training uses 105,448 clips (152.91 h), validation has 5,049 (10.77 h), and the test set has 4,610 (9.08 h). Table~\ref{tab:corpus} gives the source breakdown; Table~\ref{tab:removals} shows the clips removed during preparation. Of all clips, 43.8\% contain at least one English word, and English accounts for 24.2\% of reference words. There are 69,158 distinct words, including 14,876 written in Latin script. Durations range from 0.99 to 28.6 seconds, with a mean of 5.40 seconds.
\TableThree
\TableFour

\section{SraVaani Adaptation}
\subsection{Model architecture}
SraVaani~\cite{ref3} is a 449-million-parameter FastConformer model. Its released checkpoint was pretrained on roughly 31,000 hours of VAANI speech and then trained on a similar amount of transcribed speech. We use the released weights, vocabulary and both output heads. Table~\ref{tab:components} lists the components, and Figure~\ref{fig:training} shows the training flow.
\TableFive
\FigureThree

The front end converts 16 kHz audio to 128-dimensional log-mel features, using 25 ms windows and a 10 ms hop. Three convolutional stages reduce the frame rate by a factor of eight. A linear projection and 17 Conformer blocks~\cite{ref41} then produce encoder states of width 1024. The encoder has approximately 433 million parameters.

The main decoder is a Token-and-Duration Transducer~\cite{ref8}. A 640-unit LSTM prediction network represents the tokens emitted so far. A joint network combines that representation $g_u$ with the encoder state $h_t$:
\begin{equation}
z_{t,u}=W_o\operatorname{ReLU}(W_hh_t+W_gg_u),
\quad z_{t,u}\in\mathbb{R}^{5006}.
\end{equation}
The 5,006 output values cover 5,000 vocabulary pieces, one blank symbol and five duration choices $D=\{0,1,2,3,4\}$. Predicting durations lets the decoder advance by more than one frame. A separate CTC head predicts the vocabulary and blank at each encoder frame.

\subsection{Training objective}
We keep the checkpoint's combination of TDT and CTC losses. If $\mathbf y=(y_1,\ldots,y_U)$ is the target sequence, the TDT loss sums the probabilities of all token-and-duration paths that produce it~\cite{ref8}. With $P(v,d\mid t,u)=P_T(v\mid t,u)P_D(d\mid t,u)$, a forward variable can be written as
\begin{equation}
\begin{aligned}
\alpha(t,u)={}&\sum_{d\in D\setminus\{0\}}\alpha(t-d,u)
 P(\varnothing,d\mid t-d,u)\\
&+\sum_{d\in D}\alpha(t-d,u-1)
 P(y_u,d\mid t-d,u-1).
\end{aligned}
\end{equation}
The TDT loss is $\mathcal L_{\mathrm{TDT}}=-\log P(\mathbf y\mid\mathbf x)$. For CTC~\cite{ref39}, $\mathcal B$ collapses repeated tokens and blanks, giving
\begin{equation}
\mathcal L_{\mathrm{CTC}}=-\log\sum_{\pi\in\mathcal B^{-1}(\mathbf y)}
\prod_{t=1}^{T}P(\pi_t\mid\mathbf x).
\end{equation}
We use the original loss weighting:
\begin{equation}
\mathcal L=0.7\,\mathcal L_{\mathrm{TDT}}+0.3\,\mathcal L_{\mathrm{CTC}}.
\end{equation}
The original TDT settings for logit under-normalisation ($\sigma=0.02$) and occasional RNN-T loss use ($\omega=0.1$) are unchanged. Blanks advance by at least one frame, while token predictions may use zero duration.

\subsection{Fine-tuning configuration}
We load the released checkpoint in NeMo~\cite{ref40} and keep the tokenizer and decoder dimensions unchanged. Every training label is checked before use. Table~\ref{tab:settings} compares our settings with the official recipe.
\TableSix

The main experiment updates all model parameters. We use AdamW with $\beta_1=0.9$, $\beta_2=0.98$ and weight decay $10^{-3}$. The encoder's peak learning rate is $5\times10^{-5}$; the prediction network and output heads use $10^{-4}$. The schedule warms up for 5\% of updates, then follows cosine decay to $10^{-6}$. At update $s$, it is
\begin{equation}
\eta(s)=\begin{cases}
\eta_{\max}s/W, & s<W,\\
\eta_{\min}+\frac{\eta_{\max}-\eta_{\min}}{2}(1+\cos\theta_s),&s\ge W,
\end{cases}
\end{equation}
Here $\theta_s=\pi(s-W)/(S-W)$. For six epochs, $W=494$ and $S=9888$; for 12 epochs we use $W=989$ and $S=19776$. We clip the global gradient norm to 1 and use bf16 mixed precision.

Each mini-batch contains eight clips. We accumulate gradients over eight mini-batches, giving 64 clips per update. There are 1,648 updates per epoch. Eight workers read the tar shards, and we reshuffle shard order each epoch. Training uses the checkpoint's SpecAugment settings: two frequency masks of up to 27 bins and ten time masks of up to 5\% of the clip length~\cite{ref43}.

The TDT joint tensor can be large, especially for a long clip with many target tokens. We reduce its computation sub-batch from four clips to one. Peak reported GPU memory is 19.1 GiB. We evaluate the complete 5,049-clip validation set twice per epoch and choose the checkpoint with the lowest validation WER. The six-epoch run takes around three hours on one RTX 5090, including validation. We also train an encoder-frozen configuration, keeping the other settings the same, to test how much changing the encoder matters.

\subsection{Decoding}
We use greedy TDT decoding without an external language model. The decoder chooses a token and duration at each step, stops adding tokens for blanks and advances according to the predicted duration. We limit output to ten tokens at a frame. For the CTC comparison, we take the most likely frame-level symbols and collapse repeats and blanks.

\section{Experiments}
\subsection{Baselines and ablations}
We test released SraVaani without adaptation, full fine-tuning for six epochs, a second six-epoch seed, and a 12-epoch run. Three further runs freeze the encoder, remove numeric clips, or remove OpenSLR from training. The numeric-data ablation keeps 90,924 training clips (120.1 hours); the no-OpenSLR run keeps 36,315 clips (95.3 hours). We also measure the CTC output of the main model.

For a model-family comparison, we fine-tune Whisper large-v3-turbo (809M parameters) on the same training split. Whisper uses AdamW, a $10^{-5}$ peak learning rate, 5\% warm-up, linear decay, bf16 precision and three training epochs. The language token is Nepali, and decoding is greedy. Its checkpoint is chosen on validation WER.

We also evaluate AI4Bharat's Nepali IndicConformer. Its released 120M-parameter model has a vocabulary without Latin letters. For fine-tuning, we keep its encoder but train a new 5,000-piece BPE tokenizer and reinitialise the decoders, yielding about 128M parameters. This run uses six epochs, with peak learning rates of $10^{-4}$ for the encoder and $10^{-3}$ for the decoders.

\subsection{Test sets}
The in-domain test set contains 4,610 clips (9.08 hours) from the six sources shown in Table~\ref{tab:corpus}. We report results for the full set, for each source and for code-mixed versus Nepali-only clips. The test subset with numbers or symbols in the original transcripts has 1,125 clips. Model selection uses validation data only.

For an out-of-domain check, we use 586 clips (about 1.8 hours) from the Nepali FLEURS test split~\cite{ref34}. We exclude 138 clips containing digits because the references do not specify which spoken number form was used. The audio is unchanged, and all systems are scored on the same selected clips.

\subsection{Mixed-script FLEURS references}
FLEURS writes many English loans and names in Devanagari. Our training targets instead use Latin letters for English words. To see how much this difference affects WER, we create a second, mixed-script set of FLEURS references (examples in Table~\ref{tab:fleursrefs}). We convert English loans and foreign names to lowercase Latin script and separate attached Nepali suffixes. Common Nepali word forms, names of countries and major cities, and Nepali terms for languages and nationalities stay in Devanagari.

An LLM made the initial text conversion from the original FLEURS references without seeing model predictions. We checked phonetic consistency and conversion rules automatically, but did not complete human verification. The conversion changes 390 of 586 sentences; 9.3\% of words in the resulting references are in Latin script. This is a test of writing conventions on read speech, not a new spoken code-switching benchmark.
\TableSeven

\subsection{Evaluation metrics}\label{sec:metrics}
All scoring uses lowercased text with punctuation removed. We report corpus-level word error rate (WER) and character error rate (CER) overall and for each domain. To separate recognition from writing convention, we also use four diagnostic metrics. \emph{Language-specific WER} assigns substitution and deletion errors to the language of the reference word, and insertions by the predicted script. \emph{English exact match} measures English words written correctly in Latin script; \emph{wrong script} counts recognisable English words written in Devanagari. \emph{Romanised WER} compares approximately phonetic Latin-script forms, reducing penalties for some script and spelling differences. Finally, \emph{switch-point error} measures substitutions and deletions next to a language boundary, compared with words elsewhere~\cite{ref16,ref18,ref20,ref21}.
We use corpus-level error totals, not a mean of clip-level scores. We estimate 95\% confidence intervals with 1,000 paired bootstrap resamples over clips. These intervals describe uncertainty over clips and do not account for shared-speaker dependencies.

\subsection{Implementation details}
The experiments reported in this paper used one NVIDIA RTX 5090 with 32 GB of memory, NeMo 2.7.3 and the recorded PyTorch/CUDA environment. Training was logged with TensorBoard and Weights \& Biases.

\section{Results and Discussion}
\subsection{Main results}
SraVaani without fine-tuning reaches 48.6\% WER on the six-domain test set. After six epochs, its WER is 17.1\% (95\% CI 16.6--17.6\%). A different seed gives 17.0\%. Extending training to 12 epochs lowers WER to 16.3\%. Full results appear in Table~\ref{tab:results}. Gains are seen in all six domains, although spontaneous podcasts and news remain harder than read speech.

The two best systems have similar overall accuracy: Whisper large-v3-turbo reaches 16.4\% WER, and 12-epoch SraVaani reaches 16.3\%. SraVaani has 449M parameters compared with Whisper's 809M. The small difference between their WER scores is not statistically significant under our paired bootstrap. IndicConformer reaches 19.9\% after adaptation.
\TableEight

\subsection{Code-mixed and switch-point performance}
Table~\ref{tab:language} separates English and Nepali errors. Before fine-tuning, English-word WER is 80.4\%, and only 21.2\% of English words match the reference exactly. After six epochs, English-word WER is 20.1\% and exact recognition is 82.3\%. With 12 epochs, these numbers improve to 18.8\% and 83.6\%. Nepali-word WER also drops, from 36.1\% before adaptation to 15.9\% after six epochs.

The gains are not only a change in script. Romanised WER, which largely ignores script differences, drops from 41.3\% to 15.5\% after six epochs. Before adaptation, words next to language switches have a 65.9\% error rate compared with 35.5\% elsewhere. After adaptation, switch-point errors are 14.3\%, close to 15.2\% on other words. Code-mixed clips still have slightly higher WER than Nepali-only clips (17.4\% versus 15.2\% in the six-epoch run).
\TableNine

\subsection{Encoder adaptation and training duration}
The official fine-tuning recipe freezes the encoder. In our experiments, that gives 25.5\% WER, compared with 17.1\% when we update the full network. The difference is 8.4 WER points (95\% CI +8.0 to +8.8). The frozen-encoder system learns much of the desired output script, but its romanised WER remains 23.5\%, compared with 15.5\% for full fine-tuning. Updating the encoder therefore appears to help recognise the speech itself, not just change its written form.

The validation curves in Figure~\ref{fig:ablation-curves} show that the frozen-encoder run levels off earlier. The six-epoch full model ends at 12.94\% validation WER, while the frozen-encoder run ends near 19.9\%. Twelve epochs reduce test WER by another 0.8 points, to 16.3\%, with a 95\% CI for this change of $-1.0$ to $-0.6$. Figure~\ref{fig:training-curves} shows the two training schedules. The 12-epoch run still improves at its final checkpoint, so this experiment does not establish complete convergence.
\FigureFour
\FigureFive
\TableTen

\subsection{Effect of numeric training data}
Removing training clips whose source transcripts contain digits or symbols reduces training audio from 152.9 to 120.1 hours. This model has 19.8\% overall WER, 2.7 points worse than the main six-epoch model. The largest difference is on the 1,125 numeric test clips: WER rises from 14.3\% to 21.3\%. On the remaining test clips, the difference is small (+0.26 points; 95\% CI $-0.04$ to +0.56).

Table~\ref{tab:numbererrors} looks at number words rather than whole utterances. With the full training data, the error rates are 4.9\% for English number words and 6.4\% for Nepali number words. Without numeric training clips they rise to 16.6\% and 29.8\%. These results support keeping spoken-number examples in training, although the way we selected labels can introduce bias and needs further auditing.
\TableEleven

\subsection{Out-of-domain evaluation}\label{sec:fleurs-results}
Results on the standard Nepali FLEURS references show a loss after adaptation. SraVaani's WER increases from 22.6\% zero-shot to 31.1\% after six epochs (Table~\ref{tab:fleurs}). The two reference conventions explain part of this difference. The original references often use Devanagari for English loans, while our fine-tuned model writes those words in Latin script. With our mixed-script FLEURS references, WER instead falls from 30.0\% zero-shot to 23.4\% after six epochs.

This change does not mean the out-of-domain problem has disappeared. When we ignore both script and word boundaries, romanised CER still increases from 5.1\% to 7.6\%. Thus, part of the standard FLEURS drop is a transcription mismatch and part is a real recognition loss. The frozen-encoder model also loses accuracy, suggesting that decoder output habits contribute to the change. Training with examples from both writing conventions may be useful, but we have not tested that here.
\TableTwelve

\TableThirteen
\subsection{Effect of Nepali read speech}
OpenSLR supplies many short, mostly Nepali-only clips. Removing it makes the full-test WER 18.5\%, compared with 17.1\% for the main run. Most of the loss comes from OpenSLR test speech, where WER rises from 12.4\% to 29.3\%. On the five code-mixed domains together, WER changes by only 0.27 points, although Nepali-word errors increase slightly while English-word errors decrease slightly. Out-of-domain FLEURS WER also rises from 31.1\% to 36.2\%. In this setting, monolingual read speech helps preserve Nepali recognition and out-of-domain performance.

\subsection{Comparison with other models}
The models differ by language and speaking style. Zero-shot Whisper has a very high WER in our setup (121.5\%), partly because repeated phrases cause many insertions; 388 test outputs show such loops. Fine-tuning reduces Whisper WER to 16.4\%. It recognises English words better than 12-epoch SraVaani (14.8\% versus 18.8\% English-word WER). SraVaani does better on Nepali words (15.3\% versus 17.0\%) and on OpenSLR read speech (12.1\% versus 16.8\%).

The Whisper run used only three epochs and its validation score was still improving. We therefore do not claim that the present comparison measures the best possible Whisper configuration. Fine-tuned IndicConformer, with about 128M parameters, reaches 19.9\% WER. Much of the gap is on English words. On OpenSLR read speech, IndicConformer and SraVaani perform similarly.

\subsection{Error analysis}
Before adaptation, SraVaani often uses the wrong Indic script. It produces mainly another Indic script for 40.9\% of OpenSLR test recordings and 14.0\% of the full test set. After fine-tuning, none of the test clips is mostly in another script.

After adaptation, errors mainly involve short Nepali particles and postpositions such as \dn{को}, \dn{ले}, \dn{छ} and \dn{र} may be missed; spelling choices such as \dn{हरु} versus \dn{हरू} may differ; and uncommon English names, technical terms or programming tokens are still difficult. Some differences also come from inconsistent reference spelling. Table~\ref{tab:examples} shows representative output examples, including a difficult short code-mixed clip.

\section{Conclusion}
We adapted SraVaani to Nepali--English code-mixed speech using a 172.8-hour corpus while retaining its tokenizer and output heads. The main preparation challenge was resolving digits and symbols into labels that fit the vocabulary and the spoken language. With full fine-tuning, test WER fell from 48.6\% to 17.1\% after six epochs and to 16.3\% after 12 epochs. The encoder-frozen and data ablations show that updating the encoder and keeping numeric speech both contribute to the gain. Nepali read speech helps with Nepali-only and out-of-domain recognition.

The FLEURS experiment shows a limitation of reporting a single WER for code-mixed systems: changing the reference writing style reverses the apparent direction of the result. A script-insensitive measure still indicates some loss, so the issue is not only spelling. The next steps are to verify numeric labels independently, test speaker-disjoint splits, and train with more than one accepted transcription convention.

\section{Code, Models and Evaluation Resources}\label{sec:resources}
The project links are listed below. The GitHub repository is the main entry point for code and documentation. The three adapted model checkpoints and the test-output dataset are hosted on Hugging Face.
\begin{itemize}
\setlength{\itemsep}{1pt}
\setlength{\parsep}{0pt}
\setlength{\topsep}{3pt}
\item \textbf{GitHub code:} \url{https://github.com/AI-Lab-Pvt-Ltd/nepali-codemixed-asr}
\item \textbf{SraVaani model:} \url{https://huggingface.co/ailabpvtltd/sravaani-nepali-english-codemix-asr}
\item \textbf{Whisper model:} \url{https://huggingface.co/ailabpvtltd/whisper-large-v3-turbo-nepali-english-codemix-asr}
\item \textbf{IndicConformer model:} \url{https://huggingface.co/ailabpvtltd/indicconformer-nepali-english-codemix-asr}
\item \textbf{Test outputs and FLEURS references:} \url{https://huggingface.co/datasets/ailabpvtltd/nepali-english-codemix-asr-test-outputs}
\end{itemize}
These releases do not include the collected training audio, which remains subject to the original owners' rights and permissions.

\section*{Limitations and Ethical Considerations}
The inherited train, validation and test splits are not guaranteed to be speaker-disjoint. As a result, in-domain WER may be optimistic. FLEURS covers read speech with limited English content and does not represent spontaneous out-of-domain code-mixing. Its mixed-script references were generated with LLM assistance and automatically checked, but not independently validated by listening to each recording.

Selecting number readings from the base model's predictions may favour its existing errors. An initial audit found five incorrect labels among 150 sampled numeric clips before revising the candidate rules; the final versions still require independent review. Only the main configuration was repeated with a second training seed, and Whisper's validation WER was still improving after its shorter training schedule. Bootstrap intervals account for sampled clips, not shared speakers or run-to-run variation.

\end{document}